\documentclass[sigconf]{acmart}
\usepackage{algorithm}
\usepackage{algpseudocode}
\usepackage{booktabs}
\usepackage{multirow}
\usepackage{graphicx}
\AtBeginDocument{%
  }

\copyrightyear{2025}
\acmYear{2025}
\setcopyright{acmlicensed}
\acmConference[WWW Companion '25] {Companion of the 16th ACM/SPEC International Conference on Performance Engineering}{April 28-May 2, 2025}{Sydney, NSW, Australia.}
\acmBooktitle{Companion of the 16th ACM/SPEC International Conference on Performance Engineering (WWW Companion '25), April 28-May 2, 2025, Sydney, NSW, Australia}
\acmISBN{979-8-4007-1331-6/25/04}
\acmDOI{10.1145/3701716.3715306}
\makeatletter
\def\@copyrightpermission{
   \begin{minipage}{0.25\columnwidth}
     \href{http://creativecommons.org/licenses/by/4.0/}{%
       \includegraphics[width=\linewidth]{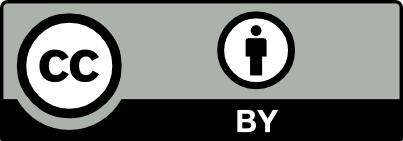}}%
   \end{minipage}\hfill
   \begin{minipage}{0.7\columnwidth}
     \href{http://creativecommons.org/licenses/by/4.0/}{%
       This work is licensed under a Creative Commons 
       Attribution 4.0 International License (CC BY 4.0).}
   \end{minipage}
   \vspace{5pt}
}
\makeatother

\begin{document}

\title{RU-AI: A Large Multimodal Dataset for Machine-Generated Content Detection}

\author{Liting Huang}
\authornotemark[1]
\email{liting.huang@student.uts.edu.au}
\affiliation{%
  \institution{University of Technology Sydney}
  \streetaddress{15 Broadway}
  \city{Sydney}
  \state{New South Wales}
  \country{Australia}
  \postcode{2000}
}

\author{Zhihao Zhang}
\authornote{Both authors contributed equally to this research.}
\email{zhihao.zhang11@unsw.edu.au}
\affiliation{%
  \institution{University of New South Wales}
  \streetaddress{1 Cleveland Street}
  \city{Sydney}
  \state{New South Wales}
  \country{Australia}
  \postcode{2000}
}

\author{Yiran Zhang}
\email{yiran.zhang1@students.mq.edu.au}
\affiliation{%
  \institution{Macquarie University}
  \streetaddress{1 Cleveland Street}
  \city{Sydney}
  \state{New South Wales}
  \country{Australia}
  \postcode{2000}
}

\author{Xiyue Zhou}
\email{xzho9805@uni.sydney.edu.au}
\affiliation{%
  \institution{The University of Sydney}
  \streetaddress{1 Cleveland Street}
  \city{Sydney}
  \state{New South Wales}
  \country{Australia}
  \postcode{2000}
}

\author{Shoujin Wang}
\email{shoujin.wang@uts.edu.au}
\affiliation{%
  \institution{University of Technology Sydney}
  \streetaddress{15 Broadway}
  \city{Sydney}
  \state{New South Wales}
  \country{Australia}
  \postcode{2000}
}

\renewcommand{\shortauthors}{Liting Huang, Zhihao Zhang, Yiran Zhang, Xiyue Zhou, and Shoujin Wang}

\begin{abstract}
The recent generative AI models’ capability of creating realistic and human-like content is significantly transforming the ways in which people communicate, create and work. The machine-generated content is a double-edged sword. On one hand, it can benefit the society when used appropriately. On the other hand, it may mislead people, posing threats to the society, especially when mixed together with natural content created by humans. Hence, there is an urgent need to develop effective methods to detect machine-generated content. However, the lack of aligned multimodal datasets inhibited the development of such methods, particularly in triple-modality settings (e.g., text, image, and voice). In this paper, we introduce RU-AI, a new large-scale multimodal dataset for robust and effective detection of machine-generated content in text, image and voice. Our dataset is constructed on the basis of three large publicly available datasets: Flickr8K, COCO and Places205, by adding their corresponding AI duplicates, resulting in a total of 1,475,370 instances. In addition, we created an additional noise variant of the dataset for testing the robustness of detection models. We conducted extensive experiments with the current SOTA detection methods on our dataset. The results reveal that existing models still struggle to achieve accurate and robust detection on our dataset. We hope that this new data set can promote research in the field of machine-generated content detection, fostering the responsible use of generative AI. The source code and datasets are available at \url{https://github.com/ZhihaoZhang97/RU-AI}.
\end{abstract}


%   \includegraphics[width=\textwidth]{data-flow.png}
%   \caption{Data collection process for all datasets. It shows the text collection (left), the image collection (middle), and the voice collection (right). Text is generated through rephrasing caption text. Images are created by referencing both caption text and real images. Voices are synthesized using caption text. All collected data is then matched with their original data pairs.}
%   \label{fig:data-collection}


\maketitle

\section{Introduction}

% The recent advent of large generative AI models has significantly impacted various sectors of human society, including industry, research, and education. They are capable of producing realistic images, human-level text and natural-sound voice, which covers the major multimedia information in communication, creation and work. Deploying generative AI models appropriately can save professionals’ time for substantive matters, assist scientists in discovering new medicine or compounds, and provide students with customised feedback. However, generative AI models’ capabilities may be misused for generating discriminate or disrespectful content, spreading misinformation, and conducting target scams. These challenges indicate the pressing need for robust mechanisms and models to verify data originality, particularly for multimodal content.

The recent advent of large generative AI models has significantly impacted various sectors of human society, including industry, research, and education. Deploying generative AI models appropriately can save professionals’ time for substantive matters, assist scientists in discovering new medicine, and provide students with customised feedback. However, generative AI models’ capabilities may be misused for generating discriminate or disrespectful content, spreading misinformation, and conducting target scams. These challenges indicate the pressing need for robust mechanisms and models to verify data originality, particularly for multimodal content. Previous studies ~\cite{automated} have explored manual methods and automated approaches for detecting machine-generated content. Researchers~\cite{researcher} have found that humans can identify machine-generat-ed content, but the quality of results is largely influenced by an individual’s background, experience and preference. Additionally, manually identifying machine-generated content is labour-intensive and difficult to scale. Therefore, it is curial to explore automated approaches to detect machine-generated content across different modalities. However, the lack of aligned multimodal datasets constrains the development of such methods. The majority of existing datasets for machine-generated content detection~\cite{genimage} are constructed within a single modality, thereby neglecting valuable cross-modal information that could enhance the detection of machine-generated content. Although there exists several multimodal datasets for machine-generated content detection~\cite{FakeAVCeleb}, they primarily focus on human faces and voices, which significantly limits their scope of applicability. Only one dataset~\cite{dgm4} includes machine-generated images and text, but lacks synthesised voice. Combining voice and images with text helps to reduce ambiguity, while integrating all three modalities can minimise information loss. This approach can be effectively applied to combat fraud and disinformation.

% Previous studies ~\cite{automated} have explored manual methods and automated approaches for detecting machine-generated content. Researchers~\cite{researcher} have found that humans can identify machine-generat-ed content with substantial accuracy, but the quality of results is largely influenced by an individual’s background, experience and preference. Additionally, manually identifying machine-generated content is labour-intensive and difficult to scale, while detection across multiple modalities is inherently more challenging than within a single modality. Therefore, it is curial to explore automated approaches to detect machine-generated content across different modalities. However, the majority of existing datasets for machine-generated content detection~\cite{genimage} are constructed within a single modality, thereby neglecting valuable cross-modal information that could enhance the detection of machine-generated content. Although there exists several multimodal datasets for machine-generated content detection~\cite{FakeAVCeleb}, they primarily focus on human faces and voices, which significantly limits their scope of applicability. Only one dataset~\cite{dgm4} includes machine-generated images and text, but lacks synthesised voice. Combining voice and images with text helps to reduce ambiguity, while integrating all three modalities can minimise information loss. This approach can be effectively applied to combat fraud and disinformation.

% Hence, it is important to identify whether a piece of content is from a human or a machine.
\begin{figure*}[h]
  \centering
  \includegraphics[width=0.8\textwidth]{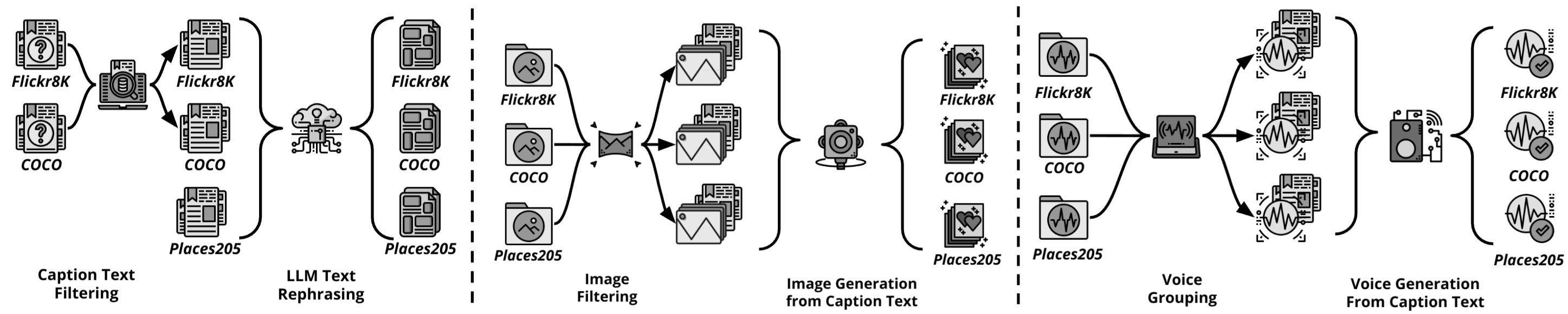}
  \vspace{-5pt}
  \caption{Data collection process for all datasets. It shows the text collection (left), the image collection (middle), and the voice collection (right). Text is generated through rephrasing caption text. Images are created by referencing both caption text and real images. Voices are synthesized using caption text. All collected data is then matched with their original data pairs.}
  \label{fig:data-collection}
  \vspace{-10pt}
\end{figure*}

% In practice, corresponding comprehensive datasets form the foundation for developing automated detection approaches and models.
% Moreover, human experts may experience difficulties in distinguishing machine-generated content from real content in specific domains, such as creative writing~\cite{10.1007/978-3-030-78635-9_67}, art painting~\cite{DBLP:conf/icccrea/ElgammalLEM17} and unfamiliar voices~\cite{DBLP:journals/corr/abs-2301-07829}.
% In addition, identifying machine-generated content manually is labour-intensive and challenging to scale, which results in machine-generated content not being identified promptly. 

To address these problems, we introduce RU-AI, a large-scale multimodal dataset for robust detection of machine-generated content in text, image and voice. The real portion of our multimodal dataset are collected from three large publicly available image caption datasets: COCO~\cite{coco}, Flickr8K~\cite{Flickr} and Places205~\cite{place} with their human voices. Machine-generated data is produced from generative AI models, including five models for text, image, and voice generation. This yields total of 1,475,370 instances each for both real and machine-generated data. We also construct a noise-augmented variant of our dataset across all modalities to evaluate model robustness. Furthermore, to demonstrate the effectiveness of our dataset, we develop and train baseline detection models on RU-AI and conduct extensive experiments to assess their performance on our dataset. Our \textbf{key contributions} can be summarised as follows:

\begin{itemize}
    % \item We introduce the first large-scale three-modal (text, image, voice) dataset, comprising 245,895 human-generated instances and 1,229,475 corresponding machine-generated instances. These machine-generated instances are produced using five current state-of-the-art (SOTA) generative models. All data instances are aligned across text, image, and voice modalities.

    \item We introduce the first large-scale three-modal (text, image, voice) dataset, comprising 245,895 human-generated instances and 1,229,475 corresponding machine-generated instances. These machine-generated instances are produced using five current state-of-the-art (SOTA) generative models.
    
    \item We introduce a variant of the proposed dataset specifically designed for model robustness testing. This version incorporates additional noise across all modalities, with three distinct types of noise applied to each modality.
    
    \item We develop baseline detection models and evaluate their performance across all modalities. Our experimental results and analysis reveal that current SOTA triple-modality models struggle to accurately verify content originality and lack robustness against noise.
\end{itemize}

\section{RU-AI Dataset}
We introduce RU-AI, a new large-scale dataset that features multimodal real and human-generated content, accompanied by machine-generated content. It encompasses three modalities: real-world images, image-describing caption text, and spoken voices for captions, providing more than 245k real and machine-generated aligned instance pairs for each modality. As shown in Figure~\ref{fig:data-collection}, image and text instances are collected from COCO~\cite{coco}, Flickr8K~\cite{Flickr} and Places205~\cite{place}, while voice instances are sourced from SLS Corpora~\footnotemark[1], which form the real and human-generated part of our dataset. The machine-generated portion is produced by five SOTA generative models specific to each modality. 

\footnotetext[1]{\url{https://sls.csail.mit.edu/}}

\subsection{Data Construction}

\textbf{Text.} Since COCO and Flickr8K incorporate five captions for each image, we extract the longest caption for each image to ensure the most comprehensive description. For Places205, which includes a single caption for each image, we sort the images' captions in descending order and extract the first 12,000 image-caption pairs. It is worth noting that Places205 has a large number of captions without punctuation, while only a small number of non-punctuated captions are present in Flickr8K and COCO. To maintain consistency across datasets, we use FullStop~\cite{fullstop} to add punctuation for all non-punctuated captions. The text data is then equally distributed among five Large Language Models (LLMs) for rephrasing. We select both open-source and proprietary LLMs based on different architectures. The rephrased generations are stored with the original captions in a structured data format identified by unique IDs.

\noindent \textbf{Image.} To ensure the quality and consistency of machine-generated images across all three datasets, we exclude images under 224 pixels in width or height and resize those over 640 pixels to a maximum of 640 pixels while maintaining their original aspect ratios. We apply five diffusion models with different parameter sizes and fine-tuning techniques to generate images, each contributing equally to the overall dataset. Image captions, positive-negative prompt pairs, and original images are used as seeds for these models during inference to ensure that the semantic meaning and realistic scenes are captured in image generation. The generated images maintain the same aspect ratio as the original images and are named after the original image IDs with the model names attached.

\noindent \textbf{Voice.} The voices for captions are cloud-sourced from a group of individuals, with each individual being assigned random captions~\footnotemark[1]. Some text-to-speech (TTS) models require reference voices for inference; therefore, we merge the spoken caption voices of each individual, identified by their unique user ID, into one-minute long voice files. For YourTTS~\cite{YourTTS}, the machine-generated voices are synthesised by using the captions along with their corresponding one-minute human voices as references. For other models, the voices are synthesised directly from the caption text using default settings. All five models contribute equally to the generated voice data, and these voices are named and saved similarly to the images, with the original IDs and model names attached. 

\noindent \textbf{Data Augmentation.} Based on the collected multimodal data, a noise-augmented variant is created towards robustness evaluation. For each modality, we introduce three different types of noise, with each type applied to 20\% of the data. As a result, the final dataset consists of 60\% for noise data and 40\% clean data. For text, the noises added at the word level are: 1) randomly removing a single character from words; 2) randomly shuffling two characters within words; 3) randomly replacing a single character in words. These alterations are applied to 20\% of the words in a sentence, and the augmented sentences adhere to the specified noise distribution. For images, three types of noise are added to randomly selected images through scikit-image~\cite{sk-image}, following the specified noise distribution: 1) Gaussian noise; 2) Salt and Pepper noise; 3) Poisson noise. For voices, three different noises are added with a 15dB Signal-to-Noise Ratio through Audiomentations~\cite{audiomentations}, following the specified noise distribution: 1) Gaussian White noise; 2) Cafe noise; 3) Rain noise.
%More details can be found in the supplementary material. 

\subsection{Data Processing}
% We employ a regular expression method to remove LLM-generated greetings, such as "Sure, here is the sentence:" and "Certainly, here's a rephrased version of your text:", from the output to maintain the quality of the text data. Additionally, as generative models may produce empty sentences or blank images due to safety and alignment-related censorship, we remove these to ensure consistency across our dataset. To ensure cross-modality alignment, we perform a data-matching process that retains the minimal overlapping subset across all three modalities. Therefore a matched data instance is formed by aligning the image, text, and voice from different modality models corresponding to a specific ID. We implement this approach in our collected dataset, resulting in 245,895 aligned real-and-AI data pairs. As shown in Table~\ref{tab:total-table}, the final processed data for each modality includes 7,890 pairs from Flickr8K, 119,264 pairs from COCO, and 118,741 pairs from Places205.

We employ a regular expression method to remove unrelated LLM-generation from the output to maintain the quality of the text data. Additionally, as generative models may produce empty sentences or blank images due to safety and alignment-related censorship, we remove these to ensure consistency across our dataset. To ensure cross-modality alignment, we perform a data-matching process that retains the minimal overlapping subset across all three modalities. Therefore a matched data instance is formed by aligning the image, text, and voice from different modality models corresponding to a specific ID. We implement this approach in our collected dataset, resulting in 245,895 aligned real-and-AI data pairs. As shown in Table~\ref{tab:total-table}, the final processed data for each modality includes 7,890, 119,264 and 118,741 pairs from Flickr8K, COCO, and Places205.

\begin{table}[]
\caption{The number of machine-generated data from different datasets in each modality of RU-AI.}
\vspace{-5pt}
\label{tab:total-table}
\scalebox{0.75}{%
\begin{tabular}{@{}ccccc@{}}
\toprule
\multirow{2}{*}{\textbf{Modality}} & \multirow{2}{*}{\textbf{Model}} & \multicolumn{3}{c}{\textbf{Dataset}}                  \\ \cmidrule(l){3-5} 
                                   &                                 & \textbf{Flickr8K} & \textbf{COCO}   & \textbf{Place}  \\ \midrule
\multirow{5}{*}{\textbf{Text}}     & ChatGPT~\footnotemark[2]                         & 1,610              & 24,272           & 23,736           \\
                                   & Llama-2-13B~\cite{llama2}                     & 1,518              & 23,121           & 23,488           \\
                                   & RWKV-5-7B~\cite{eagle}                       & 1,628              & 23,898           & 23,780           \\
                                   & Mixtral-8x7B~\cite{mixtral}                    & 1,599              & 23,752           & 23,858           \\
                                   & PaLM-2~\cite{palm2}                          & 1,535              & 24,221           & 23,879           \\ \midrule
\multirow{5}{*}{\textbf{Image}}    & StableDiffusion-1.5~\cite{stable-diffusion}             & 1,610              & 24,272           & 23,736           \\
                                   & StableDiffusion-1.5-Hyper~\cite{hypersd}       & 1,518              & 23,121           & 23,488           \\
                                   & AbsoluteReality-1.8~\footnotemark[3]              & 1,628              & 23,898           & 23,780           \\
                                   & epiCRealism-VAE~\footnotemark[4]                 & 1,599              & 23,752           & 23,858           \\
                                   & StableDiffusion-XL~\cite{sdxl}              & 1,535              & 24,221           & 23,879           \\ \midrule
\multirow{5}{*}{\textbf{Voice}}    & YourTTS~\cite{YourTTS}                         & 1,610              & 24,272           & 23,736           \\
                                   & XTTS-2\cite{xtts}                          & 1,518              & 23,121           & 23,488           \\
                                   & EfficientSpeech~\cite{efficientspeech}                 & 1,628              & 23,898           & 23,780           \\
                                   & VITS~\cite{vits}                            & 1,599              & 23,752           & 23,858           \\
                                   & StyleTTS-2~\cite{styletts}                      & 1,535              & 24,221           & 23,879           \\ \midrule
\multicolumn{2}{c}{\textbf{Total / modality}}                        & \textbf{7,890}     & \textbf{119,264} & \textbf{118,741} \\ \bottomrule
\end{tabular}%
}
\vspace{-10pt}
\end{table}

\footnotetext[2]{\url{https://openai.com/index/chatgpt/}}
\footnotetext[3]{\url{https://huggingface.co/Lykon/AbsoluteReality}}
\footnotetext[4]{\url{https://huggingface.co/emilianJR/epiCRealism}}

\subsection{Data Characteristics}

Our dataset includes 245,895 pairs of real and machine-generated data from text, image, and voice modalities. Table~\ref{tab:text-table} shows the average length of captions from different LLMs and their average similarity between the reference and generated text, indicated by average ROUGE-L and BERTScore. Table~\ref{tab:voice-table} shows the synthesised voice characteristics from different models, including average voice duration, average amplitude, and average spectral flatness. The average amplitude range refers to the average normalised audio scale, while the average spectral flatness is a measure of the sound's noise-like level. Since all images in Flickr8K are unlabelled with object instances, we employ pre-trained YOLOv5~\cite{yolov5} to annotate the unlabelled images. The data characteristics are based on the original annotations and the YOLOv5 annotations. Our dataset can be divided into 20 different categories, with "Person" being the largest category, and there are 1,040 images that do not belong to any category and are classified as "Other".

%Figure~\ref{fig:image-example} displays the object categories featured in images, which also serve as the topics in text and voice modalities.

\begin{table}[]
\caption{Statistics of generated text data. "Avg. Len." is the average number of words per sentence, "Avg. RL" is the average ROUGE-L score, and "Avg. BS" is the average BERT Score. Reference is the original text.}
\vspace{-5pt}
\label{tab:text-table}
\scalebox{0.75}{%
\begin{tabular}{@{}clccc@{}}
\toprule
\multicolumn{2}{c}{\textbf{Text Source}}       & \textbf{Avg. Len.} & \textbf{Avg. RL} & \textbf{Avg. BS} \\ \midrule
\multicolumn{2}{c}{\textbf{ChatGPT}}           & 22.97               & 0.3698           & 0.9167              \\
\multicolumn{2}{c}{\textbf{Llama-2-13B}}       & 26.96               & 0.2795           & 0.8852              \\
\multicolumn{2}{c}{\textbf{RWKV-5-7B}}         & 21.07               & 0.6411           & 0.9464              \\
\multicolumn{2}{c}{\textbf{Mixtral-8x7B}}      & 19.28               & 0.5662           & 0.9411              \\
\multicolumn{2}{c}{\textbf{PaLM-2}}            & 17.23               & 0.4615           & 0.9278              \\ \midrule
\multicolumn{2}{c}{\textbf{Reference}} & 21.66               & N/A              & N/A                 \\ \bottomrule
\end{tabular}%
}
\vspace{-5pt}
\end{table}

\begin{table}[]
\caption{Statistics of synthesised voice data. "Avg. Len." is the average duration of the voices in seconds, "Avg. AR" is the average amplitude ranges, and "Avg. BS" is the average spectral flatness. Reference is the original voices.}
\vspace{-5pt}
\label{tab:voice-table}
\scalebox{0.75}{ % Adjust the scaling factor as needed
\begin{tabular}{@{}clccc@{}}
\toprule
\multicolumn{2}{c}{\textbf{Voice Source}}       & \textbf{Avg. Len.} & \textbf{Avg. AR} & \textbf{Avg. SF} \\ \midrule
\multicolumn{2}{c}{\textbf{YourTTS}}           & 7.338                  & 1.830            & 0.1264           \\
\multicolumn{2}{c}{\textbf{XTTS-2}}            & 7.842                  & 1.826            & 0.1140           \\
\multicolumn{2}{c}{\textbf{EfficientSpeech}}   & 5.980                  & 1.416            & 0.02997          \\
\multicolumn{2}{c}{\textbf{VITS}}              & 7.536                  & 1.831            & 0.09121          \\
\multicolumn{2}{c}{\textbf{StyleTTS-2}}        & 7.563                  & 1.001            & 0.03474          \\ \midrule
\multicolumn{2}{c}{\textbf{Reference}} & 8.700                  & 1.423            & 0.03605          \\ \bottomrule
\end{tabular}
}
\end{table}

\section{Proposed Baseline}
% Based on the RU-AI dataset, we propose a unified classification model capable of determining the origin of input data from different modalities. Our model utilises a multimodal embedding model to project the features from different modalities including text, image and voice, into a common space through high-dimensional embedding sequences. These embedding sequences are then fed into a multilayer perceptron network to distinguish the origin of the input data. 

Based on the RU-AI dataset, we propose a unified classification model capable of determining the origin of input data from different modalities. Our model utilises a multimodal embedding model to project the features from different modalities into a common space through high-dimensional embedding sequences. These embedding sequences are then fed into a multilayer perceptron network to distinguish the origin of the input data. 
% \subsection{Implementation Details}

\subsection{Implementation Details} 

We use two pre-trained SOTA multimodal encoder models with their large settings as the baseline multimodal embedding models: ImageBind~\cite{ImageBind} and LanguageBind~\cite{zhu2024languagebind}. ImageBind encodes all input data into 1,024-dimensional embedding sequences, while LanguageBind encodes them into 768-dimensional embedding sequences. Each multimodal embedding model is connected to a separate multilayer perceptron network, consisting of a single 256-dimensional hidden layer. The input layers of these networks match the dimensions of their respective embedding models, 1,024-dimensional for ImageBind and 768-dimensional for LanguageBind. These networks are used for the classification of the input data as either real or machine-generated one. Specifically, the output is calculated from probability vectors $\vec{z}=(z_{r}, z_{m})$ using the softmax function $\sigma(\vec{z})_i$. The instance is classified as real if $z_r > z_{m}$; otherwise, it is classified as machine-generated. To analyse the effectiveness of different modalities in our data, we train and evaluate our classification model with different combinations of the data. The training samples are randomly selected from 80\% of each data combination, while the remaining 20\% of the samples are used as testing sets. We maintain the same size of data throughout the experiment and distribute the modalities equally. The performance is evaluated using Accuracy, Precision, and F1 metrics on both original and noise-augmented data. All models were trained for 5 epochs, using the Adam optimizer with cross-entropy loss and a learning rate of 1e-4, with the embedding model frozen. We follow the embedding models' original feature extraction techniques in all three modalities for consistency across datasets.

\begin{table}[]
%\vspace{-0.4cm}
\caption{Results of the proposed classification model with ablation studies. "LB" is LanguageBind, "IB" is ImageBind, "A" is Accuracy, "P" is Precision, "ALL" is all three modalities.}
\vspace{-5pt}
\label{tab:result-table}
\resizebox{\columnwidth}{!}{%
\begin{tabular}{@{}c|c|ccc|ccc@{}}
\toprule
\multirow{2}{*}{\textbf{Model}}                                               & \multirow{2}{*}{\textbf{Modality}} & \multicolumn{3}{c|}{\textbf{Original}}       & \multicolumn{3}{c}{\textbf{Noise Augmentation}}           \\ \cline{3-8} 
                                                                              &                                    & \textbf{A@ALL} & \textbf{P@ALL} & \textbf{F1@ALL} & \textbf{A@ALL} & \textbf{P@ALL} & \textbf{F1@ALL} \\ \midrule
\multirow{7}{*}{\textbf{\begin{tabular}[c]{@{}c@{}}LB + \\ MLP\end{tabular}}} & \textbf{Text}                      & 69.76          & 69.78          & 69.75           & 65.18          & 62.46          & 68.70           \\
                                                                              & \textbf{Image}                     & 76.49          & \textbf{85.95} & 72.93           & 62.98          & 86.95          & 45.32           \\
                                                                              & \textbf{Voice}                     & 68.89          & 64.09          & 73.43           & 68.36          & 63.66          & 73.07           \\
                                                                              & \textbf{Text+Image}                & 77.09          & 76.91          & 77.16           & 64.63          & 71.76          & 57.82           \\
                                                                              & \textbf{Text+Voice}                & 77.69          & 81.56          & 76.23           & 74.84          & 77.65          & 73.54           \\
                                                                              & \textbf{Image+Voice}               & 82.26          & 78.89          & \textbf{83.23}  & 78.20          & \textbf{91.34} & 74.12  \\ \cline{2-8} 
                                                                              & \textbf{All}                       & \textbf{84.20} & 84.20          & 82.67           & \textbf{80.02} & 78.02          & \textbf{80.74}           \\ \midrule
\multirow{7}{*}{\textbf{\begin{tabular}[c]{@{}c@{}}IB + \\ MLP\end{tabular}}} & \textbf{Text}                      & 60.70          & 57.36          & 67.96           & 64.21          & 61.42          & 68.21           \\
                                                                              & \textbf{Image}                     & 67.75          & 71.03          & 65.03           & 65.54          & 63.51          & 68.04           \\
                                                                              & \textbf{Voice}                     & 65.77          & 63.61          & 68.28           & 66.93          & 63.39          & 70.87           \\
                                                                              & \textbf{Text+Image}                & 71.48          & 75.18          & 69.23           & 68.58          & 72.41          & 65.73           \\
                                                                              & \textbf{Text+Voice}                & 75.39          & 69.05          & 78.90           & 73.05          & 67.14          & 77.06           \\
                                                                              & \textbf{Image+Voice}               & \textbf{81.25} & \textbf{82.84} & 80.78           & \textbf{79.54} & \textbf{75.99} & \textbf{80.88}           \\ \cline{2-8} 
                                                                              & \textbf{All}                       & 80.63          & 79.19          & \textbf{81.10}  & 76.42          & 72.97          & 78.11  \\ \bottomrule
\end{tabular}%
}
\vspace{-20pt}
\end{table}

% The best F1 score achieved by LanguageBind MLP is 84.20, while the best F1 score from ImageBind MLP is 81.10.
\subsection{Results}
The classification results from our proposed model are presented in Table~\ref{tab:result-table}. To evaluate the effectiveness of each modality, we conduct ablation studies using single and mixed modalities during training. In general, baseline models that utilise LanguageBind as the embedding model outperform those incorporating ImageBind. The best F1 score achieved by LanguageBind multilayer perception (MLP) is 84.20, while the best F1 score from ImageBind MLP is 81.10. This may be attributed to LanguageBind's enhanced cross-modality architecture, which binds all modalities through language, leveraging its rich semantics to provide more comprehensive information despite a smaller embedding size. For both baseline models, training with multi-modality data generally yields better performance compared to single-modality data. Since training with mixed-modality data introduces greater complexity compared to single-modality data, which could lead to better model convergence. We also observe that image and voice data tend to contribute more significantly to performance improvements. This may be caused by the richer features inherent in image and voice data compared to text. The performance trends are similar between the original data and the noise-augmented data across modalities. However, all baseline models exhibit decreased performance when evaluated on noise-augmented data, with an average drop of 2-4\%. This decrease could be attributed to the added noise disrupting data features from both human and machine sources, which can subsequently lead to a degradation in model performance during evaluation. 

Our baseline results from SOTA multimodal pre-trained models indicate that machine-generated content detection remains a challenging task, particularly for noise-augmented data. To mitigate the potential misuse of generative models, future work should focus on developing more robust and reliable detection methods.

% We observe that voice is the simplest modality to distinguish between human and machine-generated content. This aligns with the model behaviours in all-modality evaluation, as the accuracy and precision are the lowest when the model is trained only with voice data. This is likely attributable to the obvious feature difference between human and machine-generated voices. Combining data from different modalities improves the model's performance in all-modality evaluation but affects its effectiveness on the trained modality. This is expected since mixing modality data is more complex than single modality data, which could cause difficulty in model convergence. Models utilising LanguageBind as embedding models generally perform better than those incorporating ImageBind as embedding models. This might be due to LanguageBind's enhanced cross-modality architecture, which binds all modalities through language with rich semantics to provide more comprehensive information even with a smaller embedding size. According to our baseline results, machine-generated content detection remains a challenging task, especially in terms of text and cross-modal detection. To subside the potential misuse of generative models, future work is required to explore more robust and reliable methods for detecting machine-generated content.

\section{Conclusion}
We propose RU-AI, a comprehensive large-scale multimodal dataset designed for machine-generated content detection. Unlike existing datasets, which are limited to single or dual modalities, RU-AI aligns text, image, and voice data with their corresponding machine-generated counterparts. To ensure data diversity, we integrate three large open-source datasets and employ five generative models for each modality, both with and without data augmentation. Experimental results from SOTA baseline models show that unified classification models struggle to effectively detect machine-generated content on our dataset, especially when augmented with noise. This highlights the persistent challenges in the era of generative AI. By making the RU-AI dataset publicly available, we hope to foster further research in this dynamic and rapidly evolving area.

%to the public provides direction for future research. 

% \begin{acks}
% \end{acks}

\bibliographystyle{ACM-Reference-Format}
\bibliography{sample-base}

\appendix

\end{document}